\documentclass[final,5p,times,twocolumn,numbers]{elsarticle}
\usepackage{CJKutf8}

\AtBeginDocument{\begin{CJK}{UTF8}{gbsn}}
\AtEndDocument{\end{CJK}}

\usepackage{makecell}
\usepackage{amssymb}
\usepackage{rotating}
\usepackage{longtable}
\usepackage{array}
\usepackage{tabularx}
\usepackage{multirow}
\usepackage{float}
\usepackage{fvextra}
\usepackage{tcolorbox}
\tcbuselibrary{skins,breakable}
\usepackage{listings}       
\usepackage{xcolor}
\usepackage{booktabs}
\usepackage{amsmath}
\usepackage{soul, color, xcolor}

\definecolor{jsonkey}{RGB}{146, 39, 143}    
\definecolor{jsonvalue}{RGB}{58, 181, 74}   
\lstdefinelanguage{jsoncolored}{
    language=json,
    basicstyle=\ttfamily\small,
    moredelim=[s][\color{jsonvalue}]{:\ \"}{\"},
    morestring=[b][\color{jsonkey}]",
    stringstyle=\color{jsonkey},
    showstringspaces=false,
    breaklines=true,
    frame=single,
    captionpos=b
}

\lstdefinelanguage{json}{
	morestring=[b]",      
	morecomment=[l]{//},  
	stringstyle=\color{orange!70!black},
	keywordstyle=\color{blue},
	morekeywords={true,false,null},
	sensitive=false
}
\journal{Knowledge-Based Systems}

\begin{document}

\begin{frontmatter}

\title{Supervised Fine-Tuning of Large Language Models for Domain-Specific Knowledge Graph Construction: A Case Study on Hunan’s Historical Celebrities %Modern Huxiang Heroes
}

\author[a]{Junjie Hao}
\ead{haojj@hunnu.edu.cn}
\author[a,b]{Chun Wang\corref{corresponding}}
\ead{wangchun@hunnu.edu.cn}
\author[b,c]{Ying Qiao}
\ead{qy2020@hnu.edu.cn}
\author[a]{Qiuyue Zuo\corref{corresponding}}
\ead{zuoqiuyue@hunnu.edu.cn}
\author[a]{Qiya Song}
\ead{sqyunb@hnu.edu.cn}
\author[a,b]{Hua Ma}
\ead{huama@hunnu.edu.cn}
\author[a,b]{Xieping Gao}
\ead{xpgao@hunnu.edu.cn}

\cortext[corresponding]{Corresponding author：Chun Wang, Qiuyue Zuo.}

\affiliation[a]{organization={College of Information Science and Engineering},%Department and Organization
	addressline={Hunan Normal University}, 
	city={Changsha},
	postcode={410081}, 
	state={Hunan},
	country={China}}
\affiliation[b]{organization={Hunan Provincial Key Laboratory of Philosophy and Social Sciences of Yuelushan Cultural and Digital Communication (Artificial Intelligence and International Communication AIIC)},%Department and Organization
	addressline={Hunan Normal University}, 
	city={Changsha0},
	postcode={410081}, 
	state={Hunan},
	country={China}}
\affiliation[c]{organization={College of Computer Science and Electronic Engineering},%Department and Organization
	addressline={Hunan University}, 
	city={Changsha},
	postcode={410081}, 
	state={Hunan},
	country={China}}
%% Abstract
\begin{abstract}

Large language models and knowledge graphs hold broad application potential in the field of historical culture, facilitating the excavation, research, and comprehension of cultural heritage. Taking Hunan’s historical celebrities emerging from modern Huxiang culture as a case, pre-trained large models can assist researchers in rapidly extracting specific historical figure information from literature—including basic details, life events, and social relationships—and constructing structured knowledge graphs, thereby supporting related research. Currently, systematic data collection on Hunan’s historical celebrities remains scarce. Moreover, general-purpose large language models often exhibit insufficient domain knowledge extraction accuracy and weak structured output capabilities in such low-resource scenarios. Therefore, this paper proposes a supervised fine-tuning approach for domain-specific large models to enhance the quality and efficiency of information extraction regarding Hunan’s historical celebrities. Specifically, this paper first designs a fine-grained schema-guided instruction fine-tuning template for the Hunan’s historical celebrities domain. Using this template, we construct an instruction fine-tuning dataset, addressing the current lack of instruction datasets in domain-specific model fine-tuning. Second,we conducted parameter-efficient instruction fine-tuning on four publicly available large language models—Qwen2.5-7B, Qwen3-8B, DeepSeek-R1-Distill-Qwen-7B, and Llama-3.1-8B-Instruct—using the proposed instruction dataset, and established evaluation criteria for assessing their performance in character information extraction. Experimental results demonstrate that the performance of all four base models significantly improved after domain-specific fine-tuning. Among them, Qwen3-8B achieved the best performance after training with 100 samples and 50 fine-tuning iterations, scoring 89.3866 on the evaluation metrics. This research offers new insights for fine-tuning vertical large models tailored to regional historical and cultural domains, holding significant implications for promoting the cost-effective application of large models and knowledge graphs in the field of historical and cultural heritage.
\end{abstract}



%% Keywords
\begin{keyword}
Large Language Models\sep Supervised Fine-Tuning\sep Schema Definition\sep Information Extraction\sep Huxiang Culture\sep Character Knowledge Graph

%% or \MSC[2008] code \sep code (2000 is the d\alert{text}efault)

\end{keyword}

\end{frontmatter}



\section{Introduction}

%% Labels are used to cross-reference an item using \ref command.

With the rapid advancement of large language models (LLMs), unprecedented opportunities have emerged for the in-depth exploration, systematic research, and widespread dissemination of Huxiang culture. Simultaneously, this presents new challenges for the digital transformation of traditional cultural resources\cite{rf1}. Hunan’s historical celebrities serve as the core carriers of Huxiang cultural spirit. Their life stories, ideological legacy, and historical impacts form a vital component of the regional cultural heritage. Such as Zeng Guofan, Zuo Zongtang, Tan Sitong, Huang Xing,  Cai E and others shaped political reform, military modernization, cultural enlightenment, and national reconstruction through their intellectual rigor and practical action. Collectively, these celebrities embody the core attributes of Huxiang Spirit---unyielding grit and patriotism, pragmatic statecraft, and courageous innovation. Construction of a knowledge graph for this group not only provides crucial support for digital preservation of regional cultural heritage, but also lays a significant foundation for advancing the paradigm shift in intelligent historical research and enhancing cultural inheritance and dissemination.

Currently, systematic compilation of datasets on Hunan’s historical celebrities remains scarce. Existing materials are scattered across historical records, local archives, academic monographs, or isolated relational databases, lacking standardized, structured, and knowledge-based integration and annotation\cite{rf2,rf3,rf4}. This hinders direct support for intelligent research and applications. Simultaneously, general-purpose LLMs face inherent limitations when applied to such low-resource domains\cite{rf5}. Common issues include insufficient domain knowledge coverage, limited accuracy in knowledge extraction, and weak structured output capabilities.Consequently, these models perform poorly in tasks such as identifying domain-specific terminology, discerning relationships among historical celebrities, and parsing complex historical contexts. This inadequacy hinders their ability to meet the demands of in-depth exploration within the Huxiang cultural sphere.

To address the challenge of adapting general-purpose LLMs to specific domains, model fine-tuning has emerged as the core approach for domain adaptation\cite{rf6}. Among these techniques, prompt engineering offers a lightweight solution for rapid adaptation by designing targeted instructions to guide model outputs\cite{rf7,rf8}; instruction fine-tuning builds domain-specific fine-tuning datasets to enable models to develop stable capabilities on particular tasks\cite{rf9,rf10}. Meanwhile, parameter-efficient fine-tuning techniques like Low-Rank Adaptation (LoRA) achieve domain adaptation by training low-rank matrices while freezing most model parameters, significantly reducing computational resource consumption\cite{rf11}. The integration of these techniques offers a viable path for constructing domain-specific large models with low-sample and low-cost requirements. However, in scenarios like historical celebrities in modern Hunan, which combine historical specificity with data scarcity—effective adaptation of these technical approaches still requires in-depth exploration.

To this end, we focus on constructing a knowledge graph of Historical celebrities in modern Hunan, proposing a low-cost domain-specific solution based on supervised fine-tuning of LLMs. Specifically, this method utilizes open-source LLMs as a foundation, defines fine-grained schema specifications for knowledge representation, designs prompt templates tailored for historical figure knowledge extraction, and constructs domain-specific instruction fine-tuning datasets. Combining the instruction-based fine-tuning dataset with LoRA enables efficient parameter fine-tuning of the foundational large model, enhancing its knowledge extraction accuracy and structured output capabilities in resource-constrained scenarios. This provides a reusable technical pathway for constructing historical celebrities knowledge graphs.In summary, our contributions are as follows:

\begin{itemize}
	\item[(1)]  We designed an entity and relation extraction enhancement method guided by fine-grained schemas, constructing a prompt fine-tuning template for the historical celebrities domain to further enhance large language models' ability to extract historical celebrities information. Combined with the template, a prompt fine-tuning dataset specifically designed for the historical celebrities domain was constructed.
	\item[(2)]  We performed parameter-efficient instruction fine-tuning on four publicly available large language models—Qwen2.5-7B, Qwen3-8B, DeepSeek-R1-Distill-Qwen-7B, and Llama-3.1-8B-Instruct—using this instruction dataset. This optimizes the knowledge extraction capabilities of general-purpose large models in the historical figures domain and establishes domain-specific large language models. This parameter-efficient fine-tuning optimizes the knowledge extraction capabilities of general-purpose large models in the historical figures domain. We constructed a domain-specific large model focused on historical celebrities in modern Hunan, enabling supervised learning on labeled (instruction, output) pairs to infuse domain knowledge into the model. This significantly improved the accuracy of information extraction in the historical celebrities domain.
	\item[(3)]   We designed evaluation metrics for biographical information extraction. Experiments demonstrated the superiority of the fine-tuned model in the historical celebrities domain, providing a reference for intelligent research in similar historical-cultural domains with limited resources.
\end{itemize}

This paper is structured as follows: Section 1 introduces the research background, problem statement, and objectives.Section 2 reviews relevant research, including large model-assisted knowledge graph construction, domain-specific model fine-tuning techniques, and the current state of Huxiang culture digitization research. Section 3 details the proposed fine-grained knowledge schema design and domain-specific instruction dataset construction methods, explaining the LoRa-based model fine-tuning approach and knowledge extraction workflow.Section 4 validates the effectiveness of the methods through experiments and compares them with existing approaches.Section 5 summarizes the research findings, identifies limitations, and outlines future research directions.
\section{Related Work}

\subsection{Research on Large Model-Assisted Knowledge Graph Construction}
As the core carrier of structured knowledge, the core task of knowledge graph construction is to accurately extract entities, relationships and properties ( i.e., information extraction ) from unstructured text, and provide underlying support for applications such as intelligent question answering and decision support \cite{rf12}. Traditional knowledge graph construction methods rely on manual rules or supervised learning, which not only consumes a lot of labeling costs, but also has limited generalization ability when dealing with cross-domain complex semantics\cite{rf13}. With the development of LLMs such as BERT, GPT, and T5, its powerful context understanding and generation capabilities provide a breakthrough paradigm for knowledge graph construction\cite{rf14}.

The LLMs shows unique advantages in information extraction tasks.On the one hand, it can model multi-tasks such as named entity recognition, relation extraction, and triple joint extraction through a unified generative framework, without the need to design a dedicated architecture for different tasks\cite{rf15}. On the other hand, with the help of natural language prompts, data formats in different fields can be flexibly adapted to reduce the cost of task switching\cite{rf16}.For example, in the medical domain, Hu et al. utilized GPT-3.5 and GPT-4 to process complex clinical data and perform information extraction\cite{rf17}. By optimizing prompt-based extraction strategies, these models achieved substantial improvements in entity recognition tasks across multiple datasets, demonstrating their potential as efficient tools for medical information extraction. In the field of materials science, Dai et al. proposed a GPT-assisted iterative training approach, using manually annotated datasets to train knowledge extraction models, and attained an F1 score of 82.94\% in named entity recognition tasks within the domain of electromagnetic wave absorbing materials\cite{rf18}.These studies have verified the efficiency of LLMs in the construction of knowledge graphs, but their limitations in low-resource fields ( such as professional historical and cultural fields ) are still prominent. Due to the lack of domain knowledge coverage in pre-training data, the general large model is directly applied to the extraction of relational triples. Problems such as entity confusion and relationship misjudgment are difficult to meet the needs of accurate construction\cite{rf19}.

\subsection{Domain LLMs Fine-tuning Technology}
In order to solve the problem of general large model adaptation in specific fields, LLMs fine-tuning technology has become the focus of research. It has become a research hotspot to construct LLMs in vertical fields through large model fine-tuning to achieve information extraction tasks. LLMs fine-tuning techniques include prompt engineering, efficient parameter fine-tuning ( such as LoRa ), and instruction fine-tuning\cite{rf20}.

Prompt engineering guides the model output through the design domain-specific prompt template without modifying the model parameters, which is suitable for rapid verification scenarios. For example, in the field of water conservancy, Yang et al.proposed a LLMs water conservancy information extraction method based on prompt learning, which significantly improved the accuracy of entity extraction\cite{rf21}. However, this method is highly dependent on the prompt design, and the effect is unstable in complex tasks.

The parameter efficient fine-tuning technology ( such as LoRa ) preserves the fine-tuning effect while greatly reducing the computational resource requirements by freezing the model base parameters and only training the low-rank matrix. Experiments on the RoBERTa model by  Hu et al.showed that LoRa fine-tuning maintained 98 \% performance while the training parameter scale was only 1\% of the full fine-tuning\cite{rf11}.

Instruction tuning is a low-cost method based on natural language format instances to fine-tune LLMs to have domain knowledge\cite{rf22}. It organically integrates the principles of prompt engineering into LoRa fine-tuning, and fine-tunes the model by constructing a domain data set in the format of ‘instruction-input-output’, so that the model understands the domain task logic, which is the mainstream solution in low-resource scenarios.Wang et al. proposed the InstructUIE framework, which employs a unified instruction-based approach to model multiple tasks such as entity recognition and relation extraction, achieving significantly higher F1 scores on cross-domain datasets compared to traditional methods\cite{rf23}.Wang et al. ' s Self-Instruct framework automatically generates high-quality instruction samples through a small number of manual annotations, further reducing the cost of fine-tuning, and improving performance by an average of 11.5 \% in 10 domain tasks\cite{rf24}.

Although the aforementioned techniques have demonstrated effectiveness across multiple domains, systematic research remains lacking on designing tailored fine-tuning strategies for domains like modern Huxiang heroes—which possess both historical uniqueness (e.g., specific titles, associations with particular historical events) and data scarcity.

\subsection{Digital Research on Historical celebrities in Modern Hunan}
Historical celebrities in Modern Hunan, as the core carriers of Huxiang culture, possess significant historical value through their deeds and spirit. They occupy a pivotal position in the development of Huxiang culture and Chinese history, with preliminary progress achieved in related digital research. The Hunan Modern Figures Resource Database has compiled local historical records to construct a relational database containing biographical information on over 1,300 individuals, enabling structured storage of foundational data. Tianwen Digital Media has integrated its proprietary large language model with the Huxiang cultural corpus to develop the AI digital avatar “Zuo Gong”, achieving intelligent visualization of historical celebrities.

However, existing studies still face significant limitations. First, data coverage remains fragmented, with most efforts focusing on a few representative figures (e.g., Mao Zedong and Zuo Zongtang) rather than establishing a systematic collection of modern Hunan celebrities as a group. Second, the data structure is relatively homogeneous, primarily limited to basic biographical information while lacking deeper knowledge extraction such as interpersonal relationships (e.g., kinship, colleagues) and event participation. Third, the level of standardization is low—the data sources span literature, archives, and folklore, resulting in inconsistent formats and conflicting information, which hinders direct support for knowledge graph construction and domain-specific large model training. This dual challenge of “data scarcity and poor standardization” has made it difficult to effectively implement general knowledge graph technologies and domain fine-tuning methods, thus becoming a key bottleneck restricting the in-depth exploration of knowledge about historical celebrities in Modern Hunan.

In view of the above problems, we proposes a knowledge graph construction scheme for historical celebrities in Modern Hunan based on supervised fine-tuning of large language models. By defining granular schema specifications for knowledge representation, designing prompt templates tailored for domain-specific knowledge extraction, and constructing domain-specific instruction fine-tuning datasets, combined with instruction fine-tuning datasets and LoRA technology to achieve efficient parameter fine-tuning, enhancing the model's accuracy in extracting knowledge about historical celebrities in Modern Hunan and its structured output capabilities in resource-constrained scenarios. This approach enables the construction of an accurate and comprehensive knowledge graph of historical celebrities in Modern Hunan, providing robust support for the digital preservation of Huxiang cultural heritage and intelligent historical research.

\section{Methodology}
\subsection{Overall Framework }

\begin{figure*}[t] 
	\centering 
	\includegraphics[width=\linewidth]{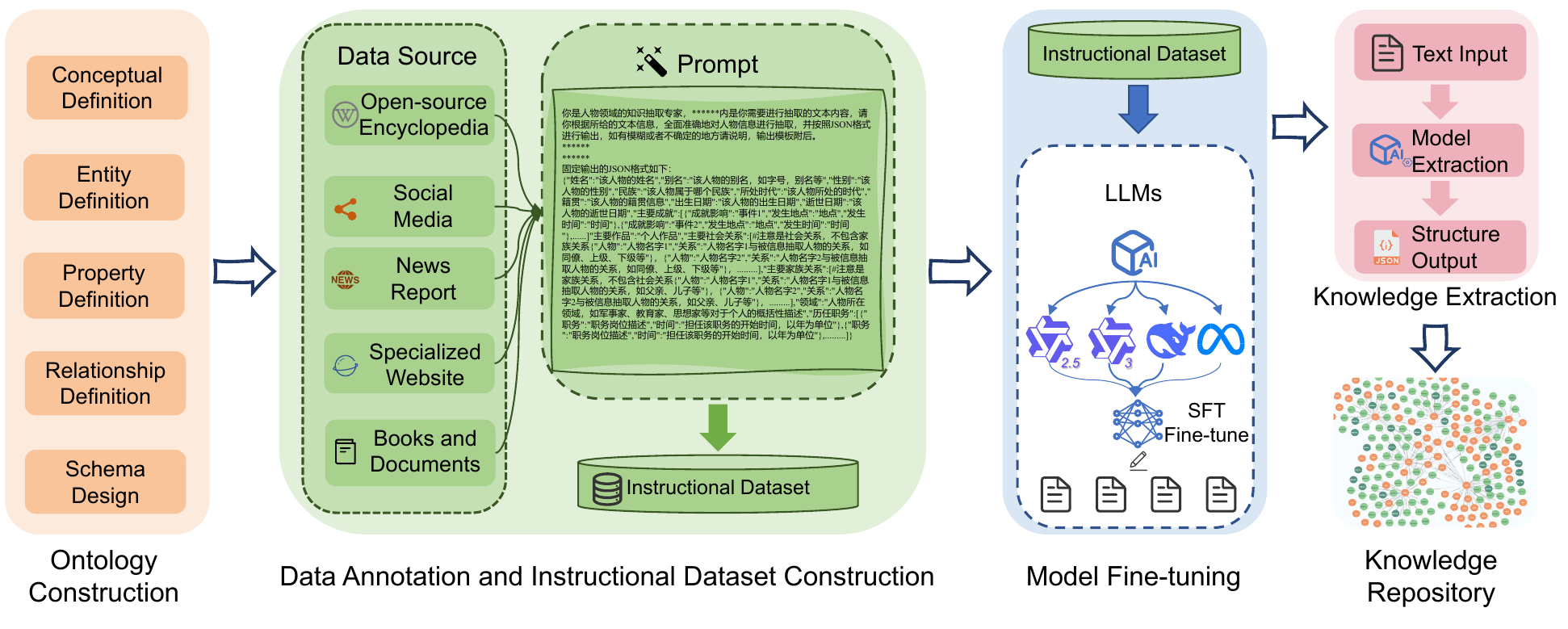} 
	\caption{Overall Research Framework} 
	\label{fig-overall} 
\end{figure*}

We propose a cost-effective method for constructing historical figure knowledge graphs by integrating LLMs with parameter-efficient fine-tuning (LoRA) technology. The overall framework, illustrated in Figure~\ref{fig-overall}, comprises four core stages: (a)ontology construction, (b)data annotation and  instructional dataset construction, (c)fine-tuning large language models and knowledge extraction, (d) knowledge graph construction. Schema provides the structural template and semantic boundaries for the entire graph construction. The fine-tuned model, combined with prompt templates, extracts structured information from unstructured text. Knowledge fusion is achieved through entity alignment and relationship normalization, with the final output imported into a graph database to enable visual queries and task applications.

\textbf{(a)Ontology Construction}: Integrating domain-specific characteristics of historical figures, this process involves literature review and expert consultation to define a fine-grained knowledge representation schema. It specifies core entity types (e.g., historical celebrities,  successive official posts,major achievements, and family relationships), entity properties (e.g., birth and death dates, courtesy names, and place of origin for individuals; occurrence time and location for events), and relationship types between entities (e.g., served under, participated in, family ties, studied under, and affiliation). This establishes a unified semantic standard for subsequent knowledge extraction. Define entity, relationship, and property types within the domain of historical celebrities in Modern Hunan, establishing a unified semantic standard for knowledge extraction to address the fragmentation of historical knowledge representation.

\textbf{(b)Data Annotation and Instructional Dataset Construction}:Constructing domain-specific fine-tuning datasets from multi-source historical materials, combining human annotation with large-model-assisted generation techniques to expand data scale, alleviating low-resource constraints, and providing high-quality training samples for model fine-tuning. Integrating data sources including encyclopedic websites, news reports, specialized website,documents and books. The instructional dataset was constructed through data cleansing (de-duplication, noise reduction, format standardization) and semi-automated annotation.

\textbf{(c)Fine-tuning Large Language Models and Knowledge Extraction}:Utilising open-source LLMs (such as Qwen, Deepseek, Llama, etc.) as base models, we employ LoRA technology for efficient parameter fine-tuning. Through supervised fine-tuning on domain-specific datasets, the model acquires knowledge extraction capabilities within the historical celebrities in Modern Hunan domain, balancing training costs with performance requirements. This specifically involves: designing prompt templates tailored to historical celebrities, specifying task instructions (e.g., ‘Extract individuals, events, and relationships from the following text, outputting structured information’) and output formats (e.g., JSON structured format); Freezing the base model's weights while training only low-rank matrix parameters (optimised through hyperparameter settings like rank and learning rate) to achieve domain knowledge injection; Conducting multi-round fine-tuning with instruction datasets to enhance the model's recognition of terminology (e.g., ‘given name’, ‘alias’, ‘courtesy name’, ‘pen name’) and implicit relationships (e.g., ‘recommended’, ‘colleague’).

\textbf{(d) Knowledge Graph Construction}:Employing a fine-tuned model to accomplish entity recognition, relation extraction, and property extraction. Through knowledge fusion, conflicts and redundant information are resolved, ultimately constructing a structured knowledge graph of historical celebrities in Modern Hunan domain. This enables visualised storage and presentation.
\subsection{Fine-grained Schema Definition of historical celebrities in Modern Hunan}
To achieve structured and unified character knowledge extraction, we extensively explores multi-source heterogeneous data from relevant encyclopedias, specialized databases, and active WeChat public accounts, focusing on the specific domain of historical celebrities in Modern Hunan. Through rigorous screening and analysis processes within this vast dataset, highly representative materials on historical celebrities in Modern Hunan were selected. Based on this foundation, the selected materials were subjected to in-depth analysis to precisely extract key entities and relationships. This enabled the design of a fine-grained schema model tailored for the knowledge graph of historical celebrities in Modern Hunan.

Table \ref{tab-schema struction} details the core components of the Schema adopted in this research. These components encompass entity types, character attributes, and a rich variety of relationship types. Regarding entity types, it defines various objectively existing individuals associated with historical celebrities in Modern Hunan. The character attributes section provides a detailed multidimensional characterization of entities. The relationship types comprehensively cover all possible logical connections between different entities. This design ensures the schema retains foundational core entity types and properties, solidifying the fundamental framework for knowledge extraction, while also maintaining the completeness of typical character relationship categories. This endows the extraction task with exceptional scalability and universality. Such characteristics not only facilitate addressing current complex and dynamic knowledge extraction demands but also provide indispensable foundational data infrastructure support for constructing subsequent domain-specific fine-tuning datasets.
\begin{table*}[h]
	\caption{Schema Structure for the Knowledge Graph of  historical celebrities in Modern Hunan} % 标题
	\label{tab-schema struction}
	\centering % 把表居中
	\begin{tabularx}{\textwidth}{>{\hsize=1\hsize}X>{\hsize=1\hsize}X} % 四个c代表该表一共四列，内容全部居中
		\toprule % 第一道横线
		Component Type & Subcategory \\
		\midrule % 第二道横线
		Entity & Person, Achievements, Works, Relationships, Positions \\
		Character  Attributes & hasName,hasAlias,hasGender, hasBirthPlace,hasEthnic, hasBirthDate,hasDeathDate,hasFiled \\
		Events Properties & Time,Location,Influence\\
		Person-Person Relationships & hasSpouse, hasParent, hasStudent, hasColleague, hasSupervisor, hasSubordinate\\
		Person-Organization Relationships & workFor, Found, belongTo\\
		Person-Events Relationships & ParticipateIn, WinAward, Create\\
		\bottomrule % 第三道横线
	\end{tabularx}
\end{table*}
A deeper analysis of the schema elements in the knowledge graph of historical celebrities in Modern Hunan reveals that they can be clearly categorized into three key components: entities, properties, and relations. Among them, entities serve as the fundamental building blocks of the knowledge graph, primarily responsible for representing objective individuals closely associated with heroic figures. These entities encompass a wide range—from personal names, which provide the most direct identifiers, to organizational institutions that reflect the social and political contexts of their activities, as well as the titles of representative works, event names, and place names. Together, they outline the temporal background and life trajectories of the celebrities from multiple perspectives.

Properties constitute an essential means of providing structured descriptions of entities. Taking events as an example, properties can accurately depict the contextual background and analyze their impact in depth, allowing the knowledge graph to present a more multidimensional representation of celebrities’ deeds and contributions.

Relations act as the connective tissue of the knowledge graph, revealing the various types of association that exist between entities. In the case of interpersonal relations, for instance, links between “person” entities may include types such as “mentor–student,” “spousal,” “classmate,” or “friendship.” The accurate identification and representation of such relationships further enrich the semantic depth of the knowledge graph, making the relational network among individuals more coherent and complete.

In summary, the schema of the knowledge graph for historical celebrities in Modern Hunan comprehensively encompasses multifaceted information such as name, alias, gender, ethnicity, historical period, ancestral origin, date of birth, date of death, major achievements, representative works, key social relationships, primary family relationships, professional fields, and official positions. This comprehensive and meticulous design aims to accurately construct the image of historical celebrities in Modern Hunan from multiple perspectives, thereby restoring their authentic historical presence and contributions.

Moreover, considering the dynamic evolution of knowledge and the changing demands of practical applications, the schema is designed with strong scalability. As research deepens and time progresses, newly discovered entities, relationships, and properties can be seamlessly integrated into the existing schema following linked data standards. This ensures the continuous updating and refinement of the knowledge graph, allowing it to maintain an accurate and comprehensive representation of the knowledge surrounding historical celebrities in Modern Hunan.

\subsection{Domain Instruction Fine-tuning Dataset Construction}
To enable the pre-trained language models(PLMs) to adapt to the knowledge extraction task within the domain of historical celebrities in Modern Hunan, a domain-specific instruction fine-tuning dataset was constructed following a three-stage process of “multi-source data collection – manual precise annotation – template-based instruction generation.” The detailed steps are as follows.

\subsubsection{Data Sources and Preprocessing}

Taking historical celebrities in Modern Hunan as an example, the dataset was constructed using multiple authoritative sources. The biographical information was primarily collected from encyclopedic websites (such as Baidu Baike, Sogou Baike, and Wikipedia), news reports (mainly from official media outlets), thematic websites (for instance, the War of Resistance Against Japan Memorial Network, which contains extensive information on Kuomintang generals) and documents. The collected materials encompass various aspects, including biographical summaries, life histories, and anecdotal stories, thereby enabling a comprehensive and multidimensional portrayal of each historical figure. This diverse and high-quality corpus provides a solid foundation for building a precise and information-rich training dataset for the task of person information extraction.

\begin{figure}[h!] 
	\centering 
	\includegraphics[width=0.9\linewidth]{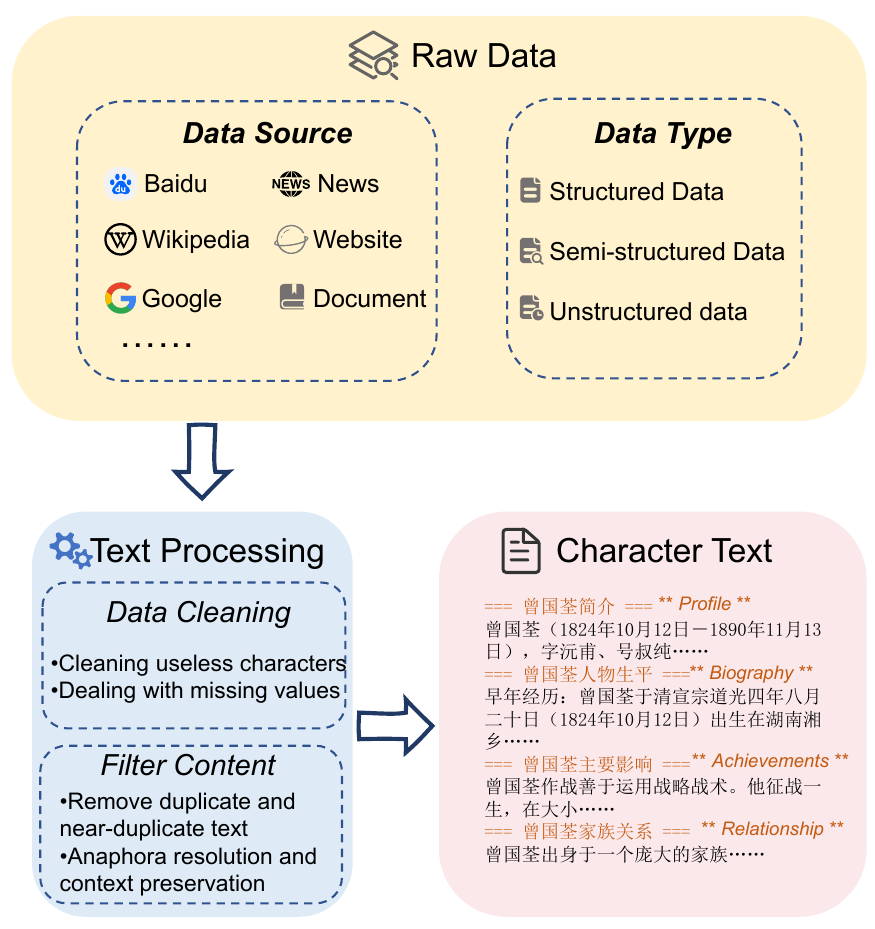} 
	\caption{The Workflow of Character Text Construction} 
	\label{fig-dataclean} 
\end{figure}

The textual data exhibit complex and diverse linguistic expressions.For instance, the presence of ambiguous words and the long-tail phenomenon common in large-scale corpora increase the difficulty of text comprehension and entity extraction. Figure~\ref{fig-dataclean} illustrates the pre-processing workflow for the textual data of historical celebrities in Modern Hunan.

First, data cleaning was performed on the raw corpus to reduce noise by removing special characters and redundant symbols, as well as detecting and eliminating duplicate or irrelevant content. Next, the texts were classified according to individual figures to ensure organized data management. To further enhance processing and analytical efficiency, each category of text was subsequently segmented into smaller units, making the corpus more structured and standardized. This pre-processing procedure effectively improved the overall quality of the textual data, providing a solid foundation for subsequent processes such as feature extraction and modeling.

\subsubsection{Prompt Template Design and  Fine-tuning Dataset Construction}
To adapt the LLMs for instruction fine-tuning and ensure close alignment with the schema structure of the knowledge graph of historical celebrities in Modern Hunan designed in this paper. We designed task-specific prompts to guide the large language model in extracting structured entity–relation information.The prompt template includes an instruction, character text, and output schema description.
Detailed prompt templates are provided in Appendix.

\textbf{Task Description (Instruction)}: The task description field provides a specific directive, clearly guiding the model on the task to be accomplished.We focus on elaborating based on the knowledge characteristics of historical celebrities in Modern Hunan and the Schema structure. In this paper, the description is formulated based on the knowledge characteristics of historical celebrities in Modern Hunan and the schema structure of the knowledge graph. The schema defines entity types, personal properties, and various relationship types, and the task description instructs the model to extract and process information in accordance with these structures.

For example: \textit{Based on the schema of the knowledge graph of historical celebrities in Modern Hunan, identify all entities in the text related to the heroes (e.g., personal names, organizational institutions, place names), delineate the types of relationships between individuals as defined in the schema (e.g., mentor–student, friendship), and extract the relevant personal properties (e.g., date of birth, major achievements).}

\textbf{Output Format}: The extracted key information should be presented in a standardized,structured format.Specifically, outputs are organized according to the unified fine-grained schema and represented in JSON.

For example, given the text:

\textit{Zeng Guofan (November 26, 1811 – March 12, 1872), courtesy name Bohán, art name Disheng, a native of Xiangxiang, Hunan (present-day Heye Town, Shuangfeng County), was known as the 'Foremost Minister of Late Qing,' a strategist, Confucian scholar, and literati. In 1853, he was commissioned to organize the Hunan militia, known as the 'Xiang Army,' whose military thought influenced generations.}

The structured JSON representation would be:

\begin{lstlisting}[language=jsoncolored, caption={Sample JSON output}]
{
    "Name": "Zeng Guofan",
    "Aliases": "Courtesy name Bohan, pseudonym Disheng",
    "Gender": "Male",
    "Era": "Mid-to-late Qing Dynasty",
    "Place of Origin": "Xiangxiang, Hunan (present-day Heye Town, Shuangfeng County, Hunan Province)",
    "Date of Birth": "November 26, 1811",
    "Date of Death": "March 12, 1872",
    "Major Achievements": {
        "influence": "Founded the Xiang Army; his military philosophy influenced subsequent generations",
        "location": "Hunan",
        "Date": "1853"
    }
}
\end{lstlisting}

This format ensures consistency, machine-readability, and alignment with the schema, downstream tasks such as model training and knowledge graph construction.

Next, the instruction dataset was constructed. Using a dataset integration approach and following the Alpaca dataset format, the preprocessed and annotated textual data of historical celebrities in Modern Hunan were combined with the prompt templates to generate instruction fine-tuning data. Each data instance was closely aligned with the schema structure, covering tasks such as entity recognition and relation extraction, thereby ensuring both task diversity and comprehensive domain coverage.

\begin{figure*}[h] %H为当前位置，!htb为忽略美学标准，htbp为浮动图形
	\centering %图片居中
	\includegraphics[width=0.9\textwidth]{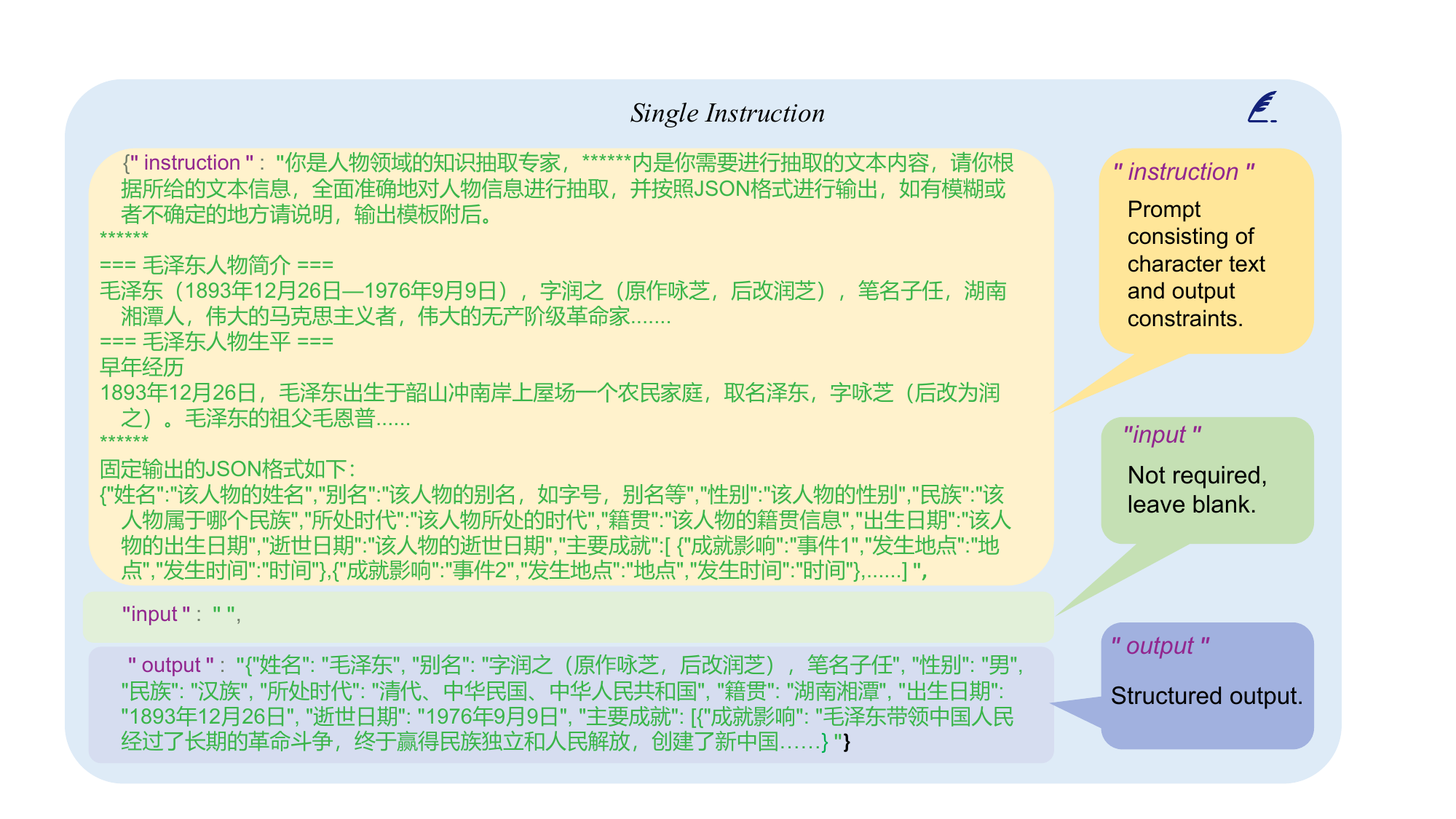} %插入图片，[]中设置图片大小，{}中是图片文件名
	\caption{Single Instruction} %最终文档中希望显示的图片标题
	\label{fig-Single Instruction} %用于文内引用的标签
\end{figure*}

In this paper, textual materials related to individual figures were concatenated with the prompt templates to serve as the input (instruction), while the manually annotated knowledge graph relations were provided in JSON format as the output. An example of a training instruction sample is illustrated in Figure \ref{fig-Single Instruction}.

By integrating the Alpaca dataset with instruction-based texts, the present study constructed an Alpaca-style instruction dataset, providing the model with a learning pathway well-suited for large language model fine-tuning. This approach emphasizes the comprehension and execution of semantically complex, instruction-driven tasks, enabling the model to more accurately capture domain-specific knowledge and the information extraction logic within the field of historical celebrities in Modern Hunan. The adoption of this integration method enhances the domain adaptability of the dataset, making it an ideal training resource for LLMs aimed at language understanding and knowledge extraction in this specialized domain.

\subsection{Supervised Fine-tuning of Large Language Models}
\subsubsection{Base Models}
We selected several currently available and high-performance LLMs as the base model for fine-tuning, including Qwen2.5-7B, Qwen3-8B, DeepSeek-R1-Distill-Qwen-7B, and Llama-3.1-8B. Considering training resource limitations and the complexity of the task, priority was given to lightweight model versions with 6B–9B parameters, ensuring both the feasibility of LoRA fine-tuning and efficient deployment.

\subsubsection{Few-Shot Fine-tuning Strategy for Base Models}

When fine-tuning the base models for tasks related to historical celebrities in Modern Hunan, we closely leveraged the previously constructed instruction fine-tuning dataset, adopting the LoRA (Low-Rank Adaptation) method while integrating a few-shot strategy to efficiently enhance the general large language models’ capability for domain-specific knowledge extraction.

Based on the instruction dataset structured according to the knowledge graph schema, high-quality samples encompassing diverse entities, relationships, properties, and events were further selected. To meet different sample size requirements (50, 100, and 150 samples), stratified random sampling was performed to ensure that each subset balanced coverage across historical periods and domains such as military, culture, and politics. For instance, the selected samples maintained a certain proportion of texts related to the military activities of Xiang Army generals (corresponding to military-related entities and relations in the schema) and the academic heritage of Hunan’s Historical Celebrities (corresponding to culture-related entities and relations), thereby ensuring comprehensive representation of domain knowledge.

LoRA fine-tuning differs from the prompt-based learning paradigm.While prompt-based learning guides a PLMs to perform tasks through carefully designed prompts without modifying the model parameters, LoRA fine-tuning freezes the main weights $W^{d\ast d}$of the pre-trained model and introduces low-rank matrices alongside the original weights to simulate parameter updates.
\begin{figure}[h] %H为当前位置，!htb为忽略美学标准，htbp为浮动图形h
	\centering %图片居中
	\includegraphics[width=0.85\linewidth]{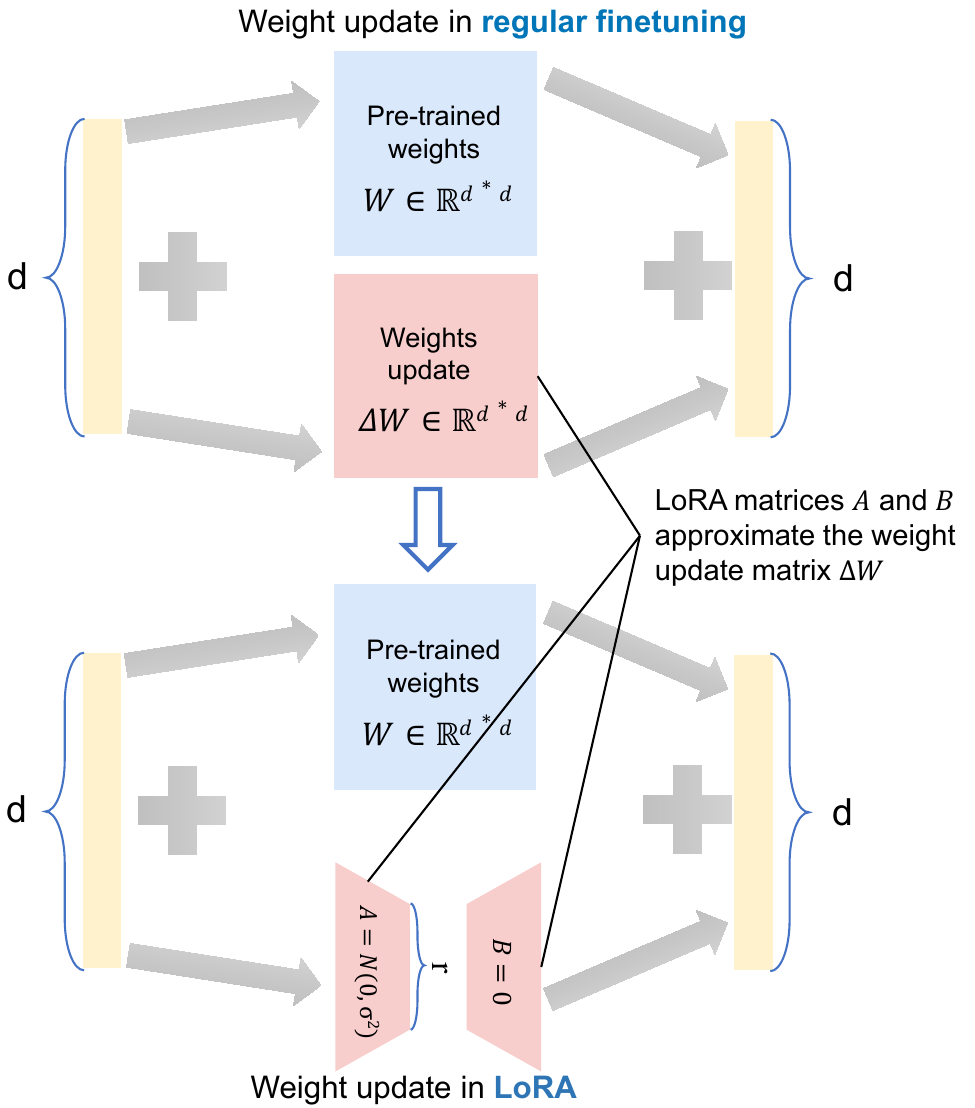} %插入图片，[]中设置图片大小，{}中是图片文件名
	\caption{Architectural Schematic of LoRA} %最终文档中希望显示的图片标题
	\label{fig-Lora} %用于文内引用的标签
\end{figure}
Specifically, during fine-tuning, only the bypass parameter matrices $A$ and $B$ are trained. At initialization, matrix 
$A$ is initialized with a Gaussian distribution, while matrix $B$ is initialized to zero, ensuring that the bypass initially has no effect on the original model (i.e., the parameter update is zero at the start of training). This architecture is shown in Figure \ref{fig-Lora}, which ignificantly reduces the number of trainable parameters for downstream tasks while preserving the original model’s comprehension and generation capabilities. Consequently, it lowers memory and computational costs, allows the fine-tuned model to achieve performance comparable to full fine-tuning, and enhances model generalization.

For each dataset corresponding to different sample sizes, LoRA fine-tuning was conducted separately. During the fine-tuning process, the prepared instruction dataset was fed into the model. Guided by the instruction, the model extracted entities, relationships, properties, and other relevant information from the text and performed contrastive learning against the provided output.

For example, given an instruction such as \textit{“Identify entities and relationships of Hunan’s historical celebrities
in the text based on the schema”}, the model learns to accurately extract schema-defined entities (e.g., \textit{“Zeng Guofan” as the entity “Huxiang heroic figure name”}) and relationships (e.g., \textit{“Zeng Guofan – led in Huxiang – Xiang Army”}).During this process, the bypass parameter matrices $A$ and $B$ are updated, thereby optimizing the model’s knowledge extraction capability within this specific domain.

\subsection{Information Extraction and KG Construction}
After fine-tuning, the large language model is capable of leveraging its knowledge extraction abilities in the domain of  Hunan’s historical celebrities to extract key information according to the schema, including:

\textbf{(a)Entity Extraction}:The model can accurately identify entities such as personal names, organizations (e.g., Huxiang-specific organizations like the Xiang Army), locations (cities, battlefields in the Huxiang region), and time points (dates of significant historical events).For example, from the sentence \textit{“Zeng Guofan organized the Xiang Army in Hunan in 1853”}, the model can correctly extract \textit{“Zeng Guofan”} (person), \textit{“1853”} (time), \textit{“Hunan”} (location), and \textit{“Xiang Army”} (organization).For ambiguous or referential expressions, the model utilizes contextual understanding and the domain knowledge acquired during fine-tuning to make accurate judgments.For instance, in a passage about Xiang Army generals stating \textit{“He led the troops to victory in the battle”}, the model can determine the specific general referred to by \textit{“he”} based on preceding context.

\textbf{(b)Relation Extraction}:The model extracts relationships between entities in accordance with the schema-defined relation types, such as \textit{“mentor–student”,“kinship/friendship”,“leadership” and “participation”}.For example, from the sentence \textit{“Zuo Zongtang was Zeng Guofan’s aide”}, the model can extract the relationship \textit{“aide (a subordinate relationship)”} between \textit{“Zuo Zongtang”} and \textit{“Zeng Guofan”}. For complex long texts, the model can analyze sentence structure and semantics to identify multiple relationships among several entities, such as hierarchical relations between generals or coordination among different units in the Xiang Army.

\textbf{(c)Property Extraction}: The fine-tuned model can accurately extract various properties of Hunan’s historical celebrities, including gender, ethnicity, major achievements, representative works, etc. For instance, from the description \textit{“Zeng Guofan, Han ethnicity, an important political figure in the late Qing, founded the Xiang Army and launched the Self-Strengthening Movement”}, the model can extract \textit{“Zeng Guofan”}’s ethnicity as \textit{“Han”} and major achievements as \textit{“founded the Xiang Army; launched the Self-Strengthening Movement”}.

\textbf{(d)Temporal Information Extraction}:Fine-tuned model precisely identifies time-related information concerning historical events in the Huxiang region, including exact dates and relative temporal references, constructing detailed and accurate chronologies of  Hunan’s historical celebrities’ lives. The extracted temporal information aligns with the schema-defined time properties.

\textbf{(e)Event Extraction}:Model accurately recognizes significant events in which the figures participated in the Huxiang region and links them with time, location, and other relevant information. The event extraction logic is fully consistent with the schema’s definitions of event-related entities and relationships.

The extracted results are imported into a graph database (e.g., Neo4j) to construct the person knowledge graph. The JSON files obtained from the large language model are first transformed and then integrated into Neo4j for fusion and visual storage. The process involves entity ID alignment and schema mapping, followed by importing data into the graph database using Cypher queries, thereby creating person nodes, relationship edges, and associated properties.

\begin{figure*}[h] %H为当前位置，!htb为忽略美学标准，htbp为浮动图形
	\centering %图片居中
	\includegraphics[width=\textwidth]{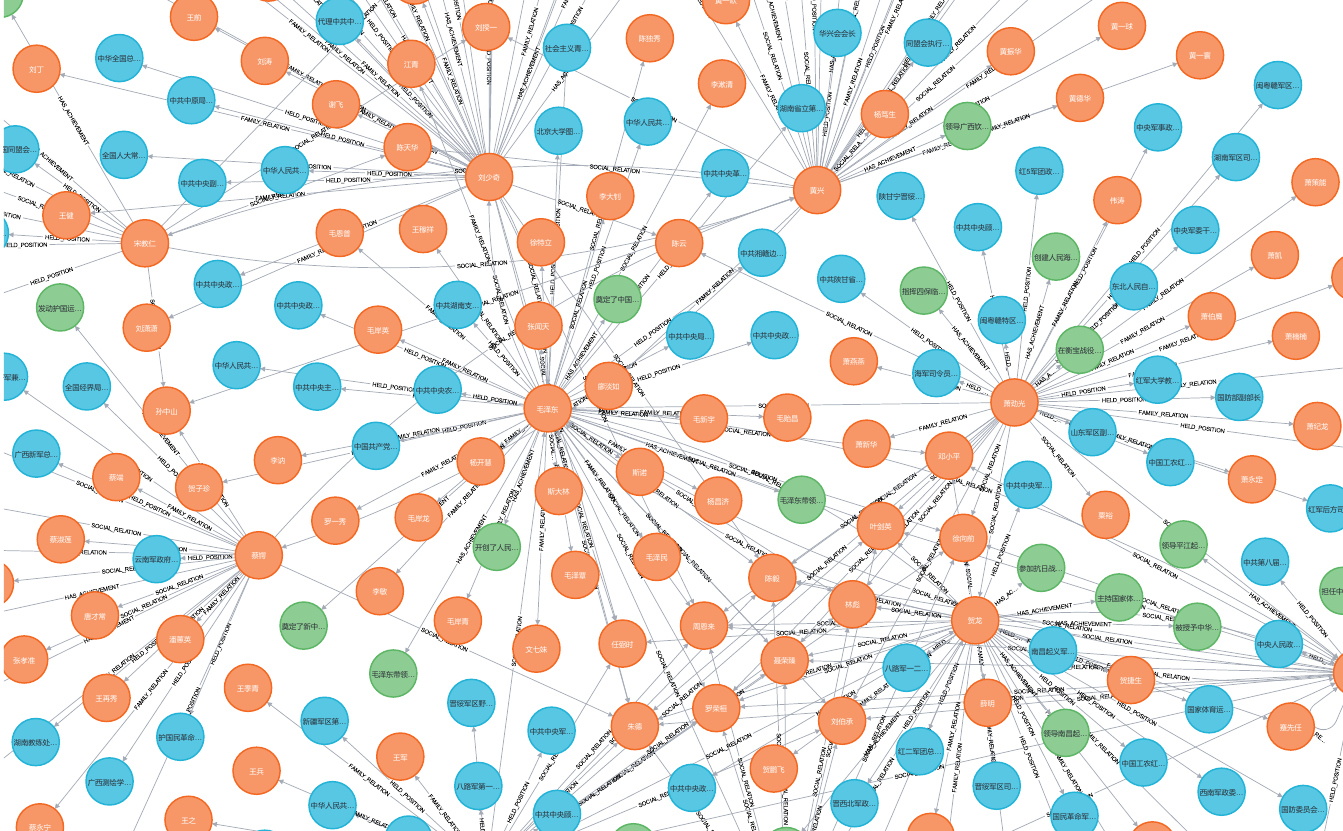} %插入图片，[]中设置图片大小，{}中是图片文件名
	\caption{View of the Knowledge Graph of Modern Huxiang Heroes} 
	\label{fig-View of KG} 
\end{figure*}

Figure \ref{fig-View of KG} illustrates the knowledge network visualization of the character knowledge graph. Through an intuitive graphical interface, users can comprehensively observe the complex relationships among individuals, the distribution of various entity types, and the interconnections between them. Different node types (e.g., person, organization, achievement) and relationship edges (e.g., \textit{“founded”,“mentor–student”,“participated”}) are distinguished using color and line style, enabling users to quickly recognize and understand the graph’s structure.

\begin{figure}[h] %H为当前位置，!htb为忽略美学标准，htbp为浮动图形
	\centering %图片居中
	\includegraphics[width=\linewidth]{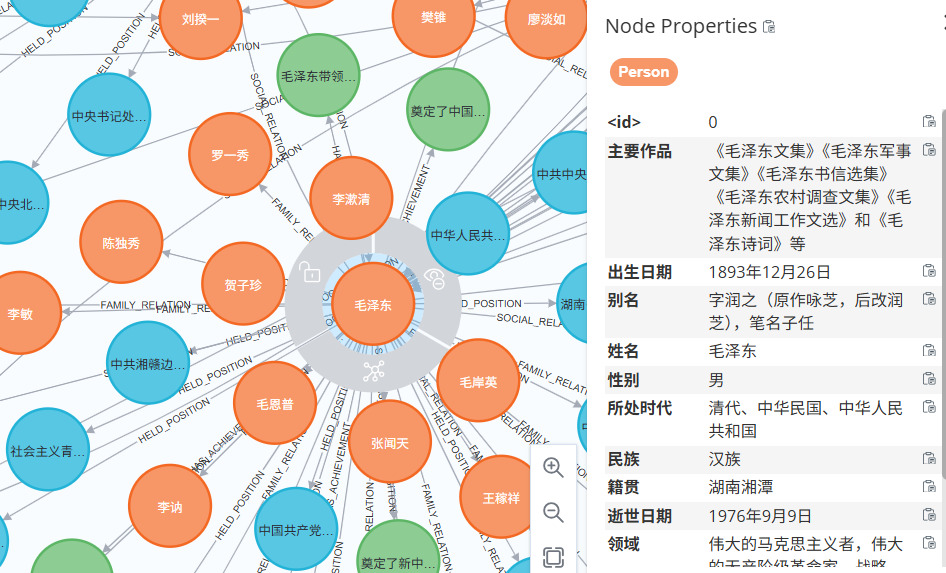} %插入图片，[]中设置图片大小，{}中是图片文件名
	\caption{Visualization of the “Mao Zedong” node in Neo4j} 
	\label{fig-MaoInKG} 
\end{figure}

Figure \ref{fig-MaoInKG} presents a local knowledge graph for a single individual (using Mao Zedong as an example), focusing on a specific person and detailing the entities and relationships directly connected to them. This localized view facilitates an in-depth study of a particular figure, illustrating their position within the knowledge graph and their connections to surrounding entities, thereby providing users with a more granular perspective of the information.

\section{Experiments}
In this section, we conduct extensive experiments to evaluate the performance of the fine-tuned large language model's ability to extract character information.
\subsection{Experiment Setting}
The experiments were conducted on an NVIDIA A800-SXM4-80GB platform. The implementation was developed in Python 3.9, based on the PyTorch 2.4.0 framework.

\subsubsection{Dataset}

The character dataset were primarily collected from open-source encyclopedic websites, official news reports, historical figure databases, and relevant scholarly literature. The dataset encompasses information such as personal profiles, life events, and anecdotes, enabling a comprehensive and multi-perspective characterization of individuals. This provides accurate and diverse linguistic resources for constructing training datasets for person information extraction.

At the initial stage, a high-quality dataset consisting of 150 manually annotated training samples and 30 testing samples was established. Both the training and testing sets include representative historical celebrities in Modern Hunan, such as Zeng Guofan (Neo-Confucian and pragmatic thinker), Mao Zedong (revolutionary spirit), and Shen Congwen (literature and art). The diversity and richness of the dataset ensure the generalization ability and robustness of the fine-tuned model.

\subsubsection{Description of the Baseline Model}

In the development of open-source LLMs, the Qwen and Llama series represent two major technological paths for Chinese and English language modeling, respectively, while DeepSeek exemplifies advances in model compression and knowledge transfer. Qwen3 and Qwen2.5 belong to the latest and previous generations of the Tongyi Qianwen family. The former demonstrates substantial improvements in language understanding, text generation, and long-text reasoning, whereas the latter excels in multilingual processing, code generation, and mathematical reasoning, serving widely as a foundational model for downstream tasks.

DeepSeek-R1-Distill-Qwen applies knowledge distillation techniques to transfer the reasoning and generation capabilities of large-scale models into smaller ones, achieving a balance between performance and efficiency—particularly suitable for scenarios with limited computational resources. In contrast, Meta’s Llama-3.1-8B-Instruct exhibits stronger generative and reasoning abilities in English contexts and benefits from a globally adopted open ecosystem.

In this paper, we employ Qwen2.5-7B, Qwen3-8B, DeepSeek-R1-Distill-Qwen-7B, and Llama-3.1-8B-Instruct as baseline models to investigate their comprehensive capabilities in person information extraction after fine-tuning.

\subsubsection{Evaluation Metric}
Traditional information extraction evaluation methods primarily rely on character-level matching, which assesses the accuracy and completeness of extraction by comparing each character of the extracted results with the corresponding characters in the ground truth\cite{rf25}. Specifically, this approach checks whether the character sequences in the extracted output exactly match those in the reference answers to determine correctness. While such metrics effectively reflect extraction performance for factual properties like name, gender, and date of birth, they are limited when applied to narrative properties such as life events or major achievements. These properties often have multiple valid expressions or variations, which traditional matching methods cannot recognize, thereby reducing the accuracy of the evaluation.

In recent years, the development of deep learning, particularly the emergence of PLMs, has significantly advanced text representation techniques. Text representation models based on PLMs have demonstrated clear superiority over traditional statistical models or shallow neural network-based approaches in both academic research and industrial applications\cite{rf26}. PLMs map variable-length texts (sentences or paragraphs) into dense vectors in a fixed-dimensional space, where similarity is typically measured using cosine similarity or dot product. In this space, semantically similar texts are represented by geometrically proximate vectors. PLMs have fundamentally transformed text vectorization and similarity assessment, as the generated sentence embeddings capture semantic information deeply, reducing misjudgments caused by differences in surface expressions.

For different types of Schema components, it is necessary to adopt matching methods of corresponding granularity, as using a single method cannot comprehensively and accurately evaluate model performance. Based on this observation, this study proposes a multi-granularity evaluation method based on Schema components. Specifically, different matching strategies are applied according to the type of Schema component: for structured information such as name, gender, date of birth, and date of death, exact matching is used; for summarized information such as major achievements and positions held, vector-space similarity is employed to assess semantic similarity with the reference answer and thus evaluate extraction performance. The final overall evaluation score is obtained by a weighted aggregation of individual scores. This multi-granularity approach produces evaluation results that more accurately reflect the model’s performance in real-world scenarios compared to a single-method evaluation.

The descriptions of each Schema component and the corresponding evaluation methods are presented in Table \ref{tab-Evaluation Rules for Schema Components}.The exact matching method compares the model-generated results with the reference answers on a character-by-character basis; a complete match receives a score of 100, while any discrepancy results in a score of 0. The vector-space similarity method employs a pre-trained large language model as a text encoder to map texts into a high-dimensional space, capturing their deep semantic relationships and thereby evaluating extraction performance. Scores range from 0 to 100, with higher values indicating greater semantic similarity. For tasks in different languages, distinct text embedding models can be used to improve evaluation accuracy. GTE-large-zh, specifically designed for Chinese texts, maps input text into a 1024-dimensional space and has achieved excellent results in Chinese multi-task vector evaluations\cite{rf27}. In this study, we adopt the general-purpose text embedding model GTE-large-zh to compute vector-space similarity for evaluation.

\begin{table*}[h]
\centering
\caption{Evaluation Rules for Schema Components}
\label{tab-Evaluation Rules for Schema Components}
\begin{tabular}{
  >{\centering\arraybackslash}p{0.0475\textwidth}
  >{\centering\arraybackslash}p{0.1425\textwidth}
  >{\centering\arraybackslash}p{0.2075\textwidth}
  >{\centering\arraybackslash}p{0.545\textwidth}
}
\toprule
No. & Component & Evaluation Method & Description and Importance \\
\midrule
1 & Name & Exact Match & Fundamental identity information, errors are unacceptable \\
2 & Alias & Vector Space Similarity & Secondary identity information, often exists in multiple versions \\
3 & Gender & Exact Match & Basic attribute, but with limited discriminative power \\
4 & Ethnicity & Exact Match & Structured attribute, moderate importance \\
5 & Era & Vector Space Similarity & Temporal contextual attribute, serves auxiliary function \\
6 & Birthplace & Vector Space Similarity & Geographical information, aids identification \\
7 & Date of Birth & Exact Match & Key timeline marker, high importance \\
8 & Date of Death & Exact Match & Key timeline marker, high importance \\
9 & Achievements & Vector Space Similarity & High semantic value, represents core significance of the figure \\
10 & Works & Vector Space Similarity & Relatively important, particularly for artists/scholars etc. \\
11 & Social Relations & Vector Space Similarity & Social network, significantly influences figure's activities \\
12 & Family Relations & Vector Space Similarity & Family network, significantly influences figure's activities \\
13 & Domain & Vector Space Similarity & Plays important role in figure positioning \\
14 & Positions Held & Vector Space Similarity & Describes career history, relatively important \\
\bottomrule
\end{tabular}
\end{table*}

The weighting of Schema components also plays a critical role in evaluation scores. To investigate how different weighting schemes affect the sensitivity of the evaluation model, three approaches were considered: equal distribution, importance-based allocation (subjective distribution), and random allocation. These schemes were used to study, under the same conditions, which weight combinations better reflect the extraction capabilities of different models. The specific weight allocation schemes are shown in Table \ref{tab-Schema Component Weight Distribution Combination}.
%origin tab3
\begin{sidewaystable}
	\centering
	\caption{Schema Component Weight Distribution Combination}
	\label{tab-Schema Component Weight Distribution Combination}
	%	\begin{tabularx}{@{}m{0.5cm} *{11}{m{1.5cm}}@{}}
		\begin{tabular}{@{}*{12}{c}@{}}
			\toprule
			%		\makebox[0.2\textwidth][c]{taskA}
			\multirow{3}*{No.} & \multirow{3}*{Component} & \multicolumn{10}{c}{Weighting Method}  \\
			\cmidrule(lr){3-12}	
			& &\multirow{2}*{\thead{Average \\ Distribution}}&\multirow{2}*{\thead{Property
					\\ Importance}}&\multicolumn{8}{c}{Random Distribution}\\
			\cmidrule(lr){5-12}	
			& & & & Random 1 & Random 2 & Random 3 & Random 4 & Random 5 & Random 6 & Random 7 & Random 8\\
			\midrule
			1 & Name & 0.07143 & 0.10 & 0.0346 & 0.1035 & 0.0997 & 0.1122 & 0.1188 & 0.0439 & 0.0373 & 0.0132\\
			2 & Alias & 0.07143 & 0.05 & 0.1038 &  0.0085 &  0.0155 &  0.0521 &  0.126 &  0.1039 &  0.1246 &  0.1447\\
			3 & Gender & 0.07143 & 0.05 & 0.0774 &  0.1038 &  0.0435 &  0.087 &  0.0890 &  0.0946 &  0.036 &  0.1108\\
			4 & Ethnic & 0.07143 & 0.05 & 0.0809 &  0.0592 &  0.033 &  0.1035 &  0.0899 &  0.1077 &  0.0728 &  0.1199\\
			5 & Era & 0.07143 & 0.06 & 0.1084 &  0.0428 &  0.099 &  0.0929 &  0.0429 &  0.0474 &  0.0398 &  0.1318\\
			6 & Birthplace & 0.07143 & 0.05 & 0.0907 &  0.0563 &  0.1118 &  0.0583 &  0.0235 &  0.0705 &  0.0369 &  0.0691\\
			7 & Date of Birth & 0.07143 & 0.07 & 0.0371 &  0.0894 &  0.02 &  0.0506 &  0.0862 &  0.0042 &  0.1323 &  0.0646\\
			8 & Date of Death & 0.07143 & 0.07 & 0.0255 &  0.0645 &  0.0035 &  0.0024 &  0.0863 &  0.0552 &  0.0011 &  0.0025\\
			9 & Achievements & 0.07143 & 0.13 & 0.1152 &  0.0796 &  0.0096 &  0.0095 &  0.1093 &  0.1096 &  0.1476 &  0.0429\\
			10 & Works & 0.07143 & 0.09 & 0.0761 &  0.1097 &  0.101 &  0.0437 &  0.0632 &  0.066 &  0.0616 &  0.0182\\
			11 & Social Relationships & 0.07143 & 0.07 & 0.0661 &  0.0684 &  0.1405 &  0.1212 &  0.0738 &  0.0679 &  0.0153 &  0.1354\\
			12 & Family Relationships & 0.07143 & 0.05 & 0.0682 &  0.0939 &  0.0396 &  0.0695 &  0.0347 &  0.0759 &  0.0799 &  0.0615\\
			13 & Domain & 0.07143 & 0.08 & 0.1056 &  0.0966 &  0.1439 &  0.0554 &  0.0393 &  0.0567 &  0.1356 &  0.0426\\
			14 & Positions Held & 0.07143 & 0.08 & 0.0104 &  0.0238 &  0.1394 &  0.1417 &  0.0171 &  0.0965 &  0.0792 &  0.0428\\
			
			\bottomrule
		\end{tabular}
\end{sidewaystable}

Using Qwen3-8B as the baseline model and fine-tuning on 100 training samples, the training batches were set to 0, 10, 30, and 50 to explore how different weight distributions impact the evaluation scores, with the results summarized in Table\ref{tab-The Impact of Different Weighting Combinations on Evaluation Model Scores}. The findings indicate that Random Combination 1 provides the most effective weight distribution for highlighting differences in model capabilities under varying fine-tuning conditions, offering a clearer and more intuitive distinction of extraction performance.

%origin tab
\begin{table}[h]
	\centering
	\caption{The Impact of Different Weighting Combinations on Evaluation Model Scores}
	\label{tab-The Impact of Different Weighting Combinations on Evaluation Model Scores}
	%	\begin{tabular}{@{}m{0.5cm} *{11}{m{1.5cm}}@{}}
		\begin{tabular}{@{}*{6}{c}@{}}
			\toprule
			%		\makebox[0.2\textwidth][c]{taskA}
			\multirow{2}*{\thead{Weighting \\ Method}} & \multicolumn{4}{c}{Fine-tune training epochs} & \multirow{2}*{Variance}   \\
			\cmidrule(lr){2-5}	
			&0 & 10 & 30 & 50&\\
			\midrule
			\thead{Average \\ Distribution} & 77.3896 & 87.3133 & 88.3156 & 88.3746 & 21.2913\\
			\thead{Property \\Importance} & 78.5486 & 86.4496 & 87.2063 & 87.4966 & 13.7000\\
			Random 1 & 73.6475 & 87.6069 & 89.0350 & 89.3866 & \textbf{42.7959}\\
			Random 2 & 79.5593 & 86.5500 & 87.0633 & 87.2276 & 10.2957\\
			Random 3 & 76.4823 & 86.9735 & 87.8421 & 88.1094 & 31.3679\\
			Random 4 & 78.1537 & 87.2814 & 88.4938 & 88.9257 & 25.8857\\
			Random 5 & 74.9264 & 86.7229 & 87.3596 & 87.5442 & 37.8387\\
			Random 6  & 79.0378 & 87.1025 & 88.0657 & 88.4379 & 19.8128\\
			Random 7 & 75.8132 & 87.5438 & 89.0124 & 89.2761 & 41.5252\\
			Random 8 & 80.3854 & 86.5249 & 88.8436 & 86.9511 & 13.4568\\
			\bottomrule
		\end{tabular}
\end{table}

Specifically, in Random Combination 1, factual properties such as name, gender, and birth/death dates are assigned lower weights, while narrative properties such as major achievements and domain of expertise are assigned higher weights. This configuration reflects the fact that LLMs already perform strongly on factual knowledge extraction, making it difficult to distinguish model capabilities based solely on these properties. In contrast, increasing the weight of narrative knowledge allows for more pronounced differentiation of models’ abilities to extract and represent person-related information under different training conditions.

\subsection{Experimental Analysis}
\subsubsection{Analysis of Fine-Tuning Effects}
Table \ref{tab-ModelPerformanceAfterFinetune} presents the performance scores of different models on the character information extraction task under a fixed training sample size of 100, with varying fine-tuning epochs. Overall, all models exhibit significant performance improvements after fine-tuning, confirming the effectiveness of the fine-tuning strategy in enhancing and transferring LLMs capabilities to specific tasks.

\begin{table*}[h]
	\centering
	\caption{Performance of the models on character information extraction after different fine-tuning epochs with a training sample size of 100.}
	\label{tab-ModelPerformanceAfterFinetune}
    \begin{tabularx}{\textwidth}{
    >{\centering\arraybackslash\hsize=2.2\hsize}X
    >{\centering\arraybackslash\hsize=0.8\hsize}X
    >{\centering\arraybackslash\hsize=0.8\hsize}X
    >{\centering\arraybackslash\hsize=0.8\hsize}X
    >{\centering\arraybackslash\hsize=0.8\hsize}X
    >{\centering\arraybackslash\hsize=0.8\hsize}X
    >{\centering\arraybackslash\hsize=0.8\hsize}X
}

			\toprule
			\multirow{2}*{Model} &  \multicolumn{6}{c}{Fine-tuning Epochs}  \\
			\cmidrule{2-7}
			& 0 & 10 & 20 & 30 & 40 & 50 \\
			\midrule
		Qwen2.5-7B & 72.0979 & 81.0805 & 85.5138 & 87.0928 & 87.5042 & 87.9128\\
		Qwen3-8B & 73.6475 & 87.6069 & 87.8262 & 89.0350 & 89.1409 & \textbf{89.3866}\\
	DeepSeek-R1-Distill-Qwen-7B& 71.2138 & 84.0245 & 85.4759 & 86.5819& 86.8913&87.8982\\
		Llama-3.1-8B-Instruct & 72.0213 & 83.0047 & 85.9012 & 87.8479 & 87.8842 & 87.9209\\
			\bottomrule
		% \end{tabularx}
    \end{tabularx}
\end{table*}

In general, model performance increases steadily with the number of fine-tuning epochs and tends to converge after approximately 30 epochs, indicating that the models have effectively learned the task patterns of character information extraction by that point. Specifically, Qwen3-8B achieves the best performance across all stages, improving from an initial score of 73.6475 to 89.3866 after 50 epochs—an overall gain of 15.74 points—demonstrating its strong ability in knowledge representation and task adaptation. Qwen2.5-7B also shows stable growth, increasing from 72.0979 to 87.9128, and its final performance is comparable to that of Llama-3.1-8B-Instruct, suggesting that it maintains robust generalization ability despite having fewer parameters.

The Llama-3.1-8B-Instruct model exhibits rapid improvement during the early fine-tuning stages (10–20 epochs), indicating that its instruction-tuning process has endowed it with transferable semantic understanding capabilities suitable for the character extraction task. However, its performance growth plateaus after 40 epochs, becoming nearly identical to that of Qwen2.5-7B. In contrast, DeepSeek-R1-Distill-Qwen-7B starts with the lowest initial score but gradually narrows the gap as fine-tuning progresses, eventually reaching 87.8982. This result highlights that, even as a distilled and lightweight model, it retains considerable potential for effective task learning through fine-tuning.

Overall, all four models demonstrate stable performance improvements through moderate fine-tuning under the condition of 100 training samples, indicating that LLMs possess strong transfer learning capabilities for character information extraction tasks even with limited supervised data. Among them, Qwen3-8B achieves the best final performance after fine-tuning, reflecting its superior semantic representation capacity and task adaptability. In contrast, DeepSeek-R1-Distill-Qwen-7B illustrates the balanced advantage of distilled models in achieving an effective trade-off between performance and computational efficiency.

\subsubsection{Comparative Experiments}

Chain-of-Thought (CoT) is a generative paradigm designed to enhance the reasoning capabilities of LLMs. Its core idea is to explicitly generate intermediate reasoning steps, decomposing complex problems into a sequence of interpretable sub-steps, thereby improving the model’s logical consistency and problem-solving accuracy. Unlike conventional direct-answer generation, CoT models “think” before they “answer”: they first construct a reasoning trajectory composed of multiple inferential links and then produce the final response based on that trajectory.

Qwen3-8B is a member of the Qwen family that supports an explicit “think / no-think” mode switch. Task prompts can activate its “thinking mode,” encouraging the model to generate chained reasoning steps incrementally, which helps it better handle complex tasks such as mathematical reasoning, code generation, and long-range logical inference.

To investigate the impact of chain-of-thought on character information extraction, this study designed a controlled experiment comparing Qwen3-8B’s extraction performance with the CoT feature enabled versus disabled under a fixed training sample size of 100, as reported in Table \ref{tab-CoTImpact}.

\begin{table}[H]
\centering
\caption{Evaluation scores of enabling or disabling the model’s chain-of-thought function on its information extraction capability.}
\label{tab-CoTImpact}
\begin{tabular}{@{}*{7}{>{\centering\arraybackslash}p{\dimexpr(\columnwidth-14\tabcolsep)/7\relax}}@{}}

\toprule
\multirow{2}*{CoT} & \multicolumn{6}{c}{Fine-tuned Epochs} \\
\cmidrule{2-7}
 & 0 & 10 & 20 & 30 & 40 & 50 \\
\midrule
Enable & 71.7349 & 84.2653 & 84.6247 & 84.8230 & 84.4835 & 85.1048\\
Disable & 73.6475 & 87.6069 & 87.8262 & 89.0350 & 89.1409 & 89.3866\\
\bottomrule
\end{tabular}
\end{table}

The experimental results indicate that enabling the Chain-of-Thought (CoT) mechanism not only fails to improve accuracy on the character information extraction task but actually degrades it. During the experiments we observed the following issues: (a) Enabling CoT causes a severe decline in computational efficiency. Because the model generates lengthy intermediate reasoning steps, it consumes far more compute and time than direct extraction, creating an untenable throughput bottleneck in large-scale text-processing scenarios. (b) Introducing CoT substantially increases the risk of factual “hallucinations” and reasoning errors. Information extraction requires fidelity to the source text, yet the model’s internal reasoning often performs “over-reasoning” based on its implicit knowledge, producing entity relations or properties that do not exist in the original documents and thereby undermining result accuracy and reliability. (c) CoT reduces the robustness of model outputs. When the model emits its internal chain as part of the output, the longer generated text exacerbates contextual memory decay and concentrates probability mass on high-probability token sequences, which in turn encourages repetitive or looping output patterns.

In summary, although Chain-of-Thought is an important technique for enhancing complex reasoning in large models, its application to character information extraction exhibits clear limitations. The core problem stems from a mismatch between the intrinsic characteristics of the task and the properties of the technique: character information extraction is fundamentally a high-precision, high-efficiency task centered on identification, classification, and localization, whereas CoT’s incremental, verbose reasoning—despite improving interpretability—introduces several practical drawbacks that cannot be ignored.

\subsubsection{Ablation Experiment}
\begin{figure*}[h]
	\centering %图片居中
	\includegraphics[width=0.9\textwidth]{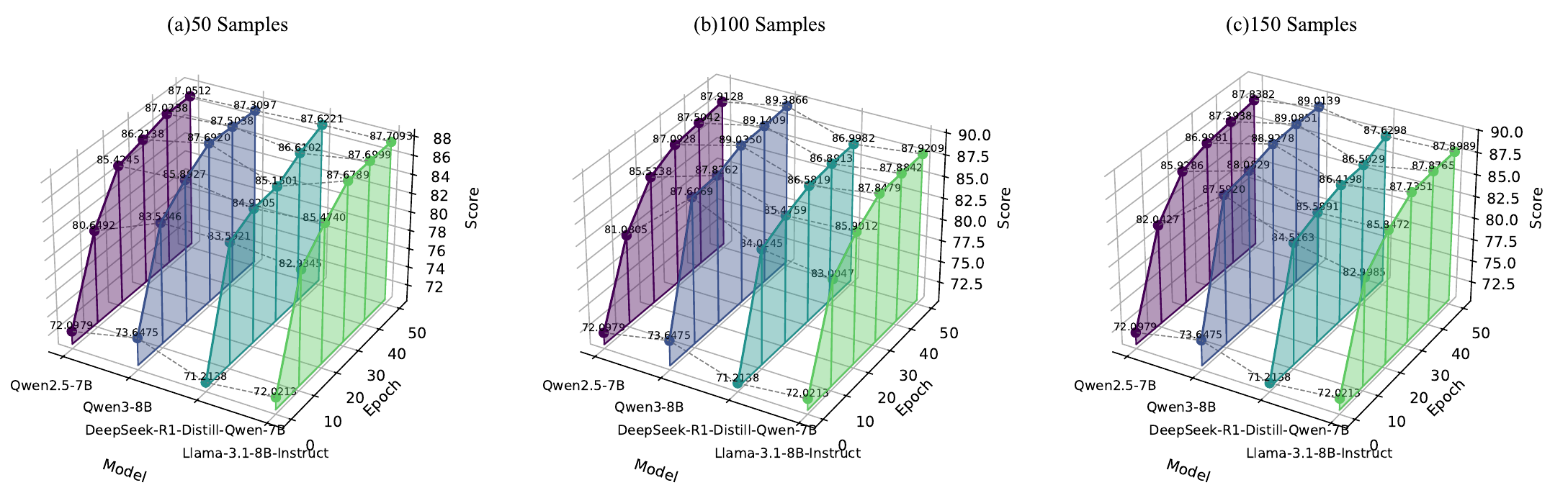} %
	\caption{Waterfall Plot: Evaluation Scores of Different Models Across Multiple Training Samples and Training Iterations}
	\label{fig-Waterfall Plot}
	
\end{figure*}

\begin{figure*}[h]
	\centering %图片居中
	\includegraphics[width=0.9\textwidth]{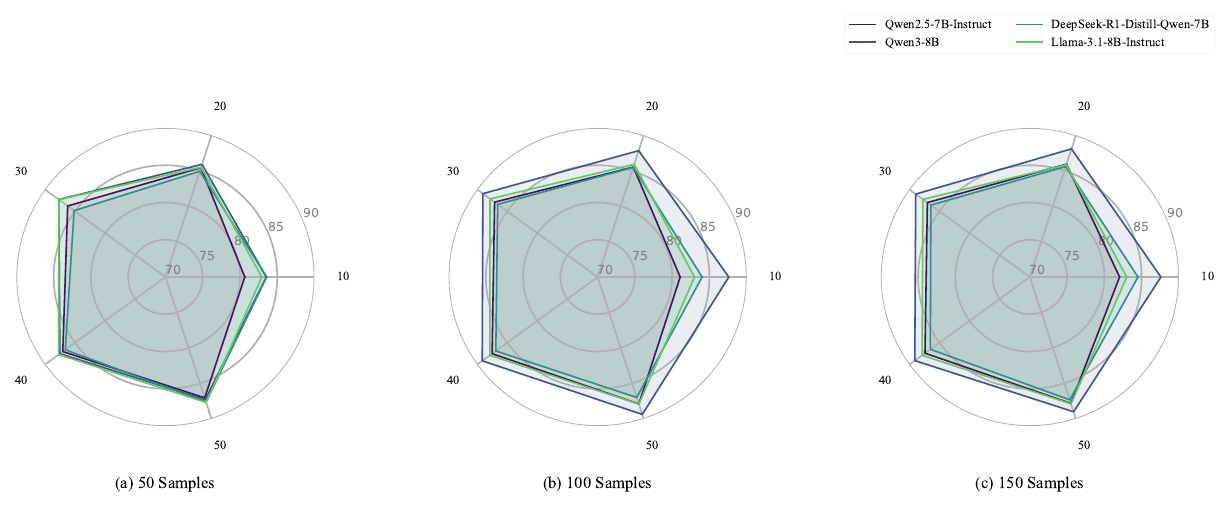} %
	\caption{Radar Plot: Evaluation Scores of Different Models Across Multiple Training Samples and Training Iterations}
	\label{fig-Radar Plot}
\end{figure*}

To comprehensively examine the effects of training sample size and fine-tuning epochs on model extraction performance, we conducted ablation experiments on the four aforementioned models under three training sample sizes (50, 100, and 150) and five fine-tuning epochs (10, 20, 30, 40, and 50). The results are presented in Table \ref{tab-Evaluation of different models across multiple training sample sizes and training epochs.}, Figure \ref{fig-Waterfall Plot}, and Figure \ref{fig-Radar Plot}.
%origin 
\begin{table*}[h]
	\centering
	\caption{Evaluation Scores of Different Models Across Multiple Training Sample Sizes and Training Epochs.}
	\label{tab-Evaluation of different models across multiple training sample sizes and training epochs.}
\begin{tabularx}{\textwidth}{
    >{\centering\arraybackslash\hsize=1.6\hsize}X
    >{\centering\arraybackslash\hsize=0.8\hsize}X
    >{\centering\arraybackslash\hsize=0.9\hsize}X
    >{\centering\arraybackslash\hsize=0.9\hsize}X
    >{\centering\arraybackslash\hsize=0.9\hsize}X
    >{\centering\arraybackslash\hsize=0.9\hsize}X
}
		\toprule
		\multirow{2}*{Model} & \multirow{2}*{Initial} & \multirow{2}*{Epochs} & \multicolumn{3}{c}{Sample size}\\
		\cmidrule{4-6}
		& & & 50 &100&150\\
		\midrule
		\multirow{5}*{Qwen2.5-7B}&\multirow{5}*{72.0979}&10&80.6492&81.0805&82.0427\\
		                                               &&20&85.4245&85.5138&85.9286\\
		                                               &&30&86.2138&87.0928&86.9981\\
		                                               &&40&87.0238&87.5042&87.3938\\
		                                               &&50&87.0512&\textbf{87.9128}&87.8382\\
		\midrule
		\multirow{5}*{Qwen3-8B}&\multirow{5}*{73.6475}&10&83.5346 & 87.6069 & 87.5920\\
													&&20 & 85.8927 & 87.8262 & 88.0829\\
													&&30 & 87.6920 & 89.0350 & 88.9278\\
													&&40 & 87.5038 & 89.1409 & 89.0851\\
													&&50 & 87.3097 & \textbf{89.3866} & 89.0139\\
		\midrule
		\multirow{5}*{DeepSeek-R1-Distill-Qwen-7B}&\multirow{5}*{71.2138}&10 & 83.5921 & 84.0245 & 84.5163\\
		&&20 & 84.9205 & 85.4759 & 85.5891\\
		&&30 & 85.1501 & 86.5819 & 86.4198\\
		&&40 & 86.6102 & 86.8913 & 86.5029\\
		&&50 & 87.6221 & 86.9982 & \textbf{87.3298}\\
		\midrule
		\multirow{5}*{Llama-3.1-8B-Instruct}&\multirow{5}*{72.0213}&10 & 82.9345 & 83.0047 & 82.9985\\
		&&20 & 85.4740 & 85.9012 & 85.8472\\
		&&30 & 87.6789 & 87.8479 & 87.7351\\
		&&40 & 87.6999 & 87.8842 & 87.8765\\
		&&50 & 87.7093 & \textbf{87.9209} & 87.8989\\
		\bottomrule
	\end{tabularx}
\end{table*}

The experimental results reveal a clear and consistent pattern in model performance across different training epochs and sample sizes. Without fine-tuning, the performance differences among the four models on the character information extraction task are relatively minor. 

However, as the number of fine-tuning epochs increases, all models exhibit substantial performance improvements, particularly during the early training stages. For instance, Qwen2.5-7B improves from an initial score of 72.0979 to 80.6492–82.0427 after 10 fine-tuning epochs, and further reaches 85.4245–85.9286 after 20 epochs. Beyond 30 epochs, the rate of improvement slows down considerably, indicating a convergence trend in model learning.

The impact of sample size on model performance varies with the number of training epochs. In the early training phase, larger sample sizes (100 or 150) contribute significantly to score improvements. For example, after 10 epochs of fine-tuning, Qwen3-8B achieves a score of 83.5346 with 50 samples, whereas it reaches 87.6069 with 100 samples—showing a substantial performance gain from increased data. However, in later training stages (40–50 epochs), the effect of sample size on final performance becomes less pronounced. Some models achieve comparable or only slightly higher scores with larger sample sizes, suggesting that as training progresses, the model’s sensitivity to sample quantity diminishes.

Different models also exhibit varying sensitivities to sample size. Qwen3-8B and DeepSeek-R1-Distill-Qwen-7B are more sensitive to sample size variations during early training, showing rapid performance gains with more data, while Qwen2.5-7B and Llama-3.1-8B-Instruct maintain more stable performance across different sample sizes, indicating stronger robustness to data fluctuations.

In terms of model comparison, Qwen3-8B achieves the best overall performance under large-sample and high-epoch conditions (scoring 89.3866 with 50 training epochs and 100 samples), indicating strong generalization ability when sufficient data and training iterations are provided. Qwen2.5-7B and Llama-3.1-8B-Instruct reach comparable final performance levels, slightly below that of Qwen3-8B. DeepSeek-R1-Distill-Qwen-7B shows rapid improvement in the early training stages but ultimately attains the lowest final score, suggesting a faster learning rate but weaker convergence capability.

Overall, all models exhibit a characteristic “rapid rise–gradual convergence” performance curve, with both training epochs and sample size exerting significant influence on extraction capability. The experimental results demonstrate that LLMs can achieve substantial performance gains on the character information extraction task through small-sample supervised fine-tuning, offering an efficient pathway for constructing structured knowledge graphs in low-resource scenarios.

\subsubsection{Case Study}
\begin{figure}[H]
	\centering %图片居中
	\includegraphics[width=\linewidth]{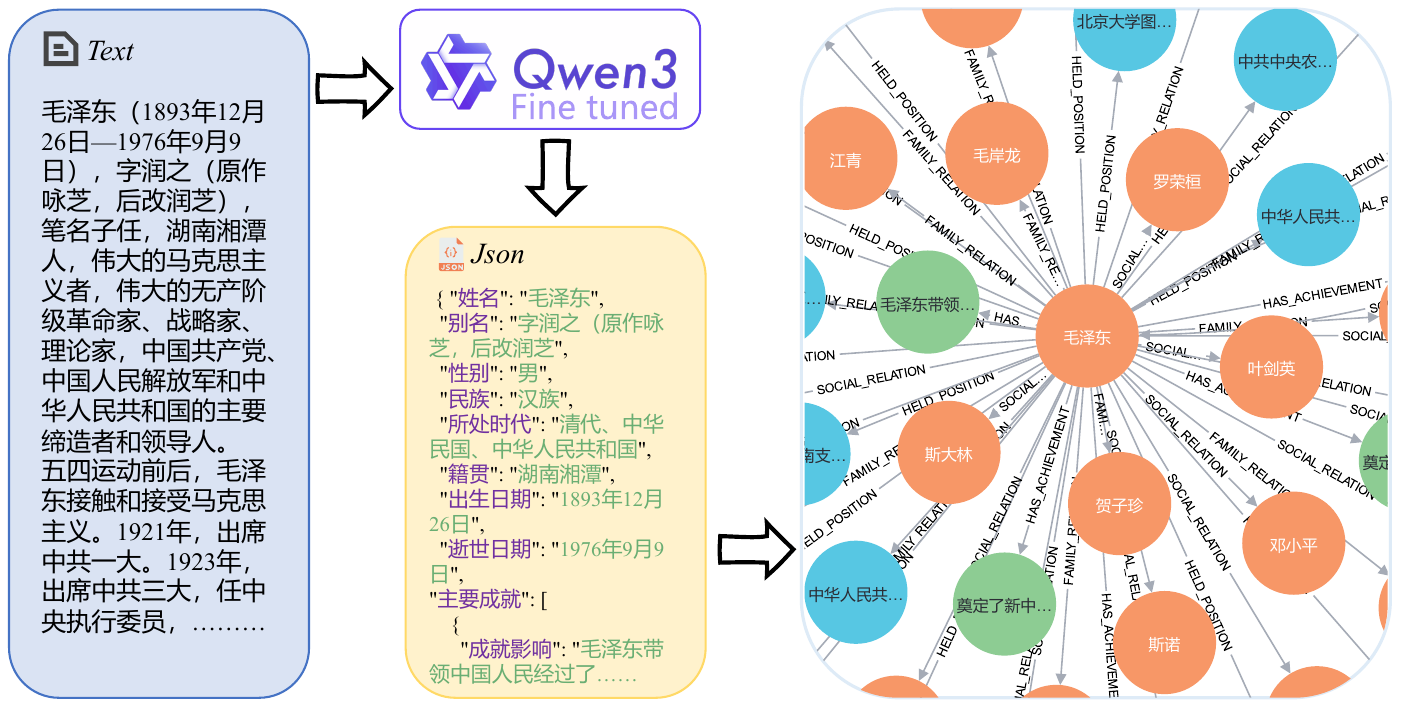} %
	\caption{Case:Taking ‘Mao Zedong’ as an example—The process of automatically constructing a knowledge graph}
	\label{fig-case}
\end{figure}

To further validate the feasibility and effectiveness of the proposed character information extraction and knowledge graph construction framework, we present a case study using Mao Zedong, one of the representative historical celebrities in Modern Hunan. The input text was selected from publicly available encyclopedic sources. The fine-tuned LLMs was employed as the information extraction module to automatically identify and structurally represent key elements, including the individual’s basic information, major achievements, social relationships, family relationships, and career experiences. Figure \ref{fig-case} illustrates the workflow from character text processing to knowledge graph construction.
From the visualization results, the knowledge structure generated by the fine-tuned model exhibits clear hierarchical organization. The semantic relationships among the character node, associated events, locations, and organizations are well-defined, indicating that the model effectively captures key entities and their semantic associations within the text. These findings confirm the capability of LLMs to perform reliable entity recognition and relation extraction from multi-source unstructured textual data, thereby supporting the construction of comprehensive and interpretable knowledge graphs.

\subsubsection{Discussion}
The supervised fine-tuning approach for LLMs leverages their extensive pretraining knowledge to deeply model complex semantic relationships, latent properties, and contextual cues within character texts. Compared to traditional extraction methods based on rules or shallow models, LLMs demonstrate superior semantic generalization and cross-domain transfer capabilities, enabling them to effectively process diverse text sources such as historical archives, biographies, and news reports. Moreover, rather than training a dedicated LLMs for character information extraction from scratch, supervised fine-tuning allows the model to quickly adapt to domain-specific extraction tasks using relatively few high-quality annotated samples, achieving “few-shot” or even “zero-shot” extraction. This approach eliminates heavy reliance on manually crafted rules or feature engineering, offering strong scalability and flexibility. In addition, fine-tuned models can not only recognize entities and properties but also produce structured outputs, automatically mapping unstructured text into knowledge graph nodes and relationships. This significantly enhances both the automation of character knowledge extraction and the efficiency of knowledge graph construction.

Despite their strong capabilities in character information extraction and knowledge graph construction, LLMs still face limitations in identifying implicit or ambiguous information in a person’s life, such as indirect relationships, temporal evolution, or semantic metaphors. The accuracy of current supervised fine-tuning methods remains limited in these contexts, particularly when distinguishing between multiple entities or co-occurring events, where the model may confuse or omit relevant information.

Future research could explore how LLMs-constructed knowledge graphs can, in turn, enhance models’ own semantic understanding, reasoning, and generative abilities, forming a closed-loop system of “knowledge extraction—graph construction—model enhancement.” This direction not only helps improve the knowledge consistency and factual accuracy of LLMs but also provides a theoretical and methodological foundation for developing intelligent systems capable of continual learning and knowledge evolution.

\section{Conclusion}
In conclusion,this study proposes a few-shot supervised fine-tuning approach for large language models, specifically designed for extracting information on historical celebrities in Modern Hunan, aiming to provide guidance for efficiently constructing knowledge graphs in low-resource, cost-constrained environments. The method leverages supervised fine-tuning to substantially enhance the performance of large models on character information extraction tasks. By defining a fine-grained schema to constrain model outputs, it enables efficient construction of historical celebrities knowledge graphs even under limited resource conditions.

Additionally, this work introduces an evaluation methodology tailored to assessing large models’ extraction of structured character information. Compared to traditional information extraction evaluation methods, this approach offers a more comprehensive and accurate reflection of model performance. Experimental results demonstrate significant improvements in extraction capabilities for fine-tuned LLMs relative to their untrained counterparts. Among the tested configurations, Qwen3-8B with 100 training samples and 50 fine-tuning epochs achieved the best performance, providing a practical and cost-effective solution for constructing historical character knowledge graphs.

Despite these advancements, the paper has certain limitations. Future research could explore more complex schemas, such as event chains or relationship chains, and investigate the integration of multi-modal information. Such directions are expected to further enhance the effective application of large language models and knowledge graphs in the domain of historical and cultural studies.

\section*{Acknowledgment}
This work is supported in part by the National Natural Science Foundation of China under Grant 62406109, the Natural Science Foundation of Hunan Province under grants 2024JJ6320, and the Hunan Provincial Educational Committee Foundation under grants 24C0005.
% \newpage
\appendix

\section*{Appendix A. Prompt Template for Character Information Extraction}
\addcontentsline{toc}{section}{Appendix A. Prompt Template for Character Information Extraction}
\subsection*{A.1 Task Description}
The following prompt was designed for character-centric knowledge extraction tasks.  
The large language model is instructed to comprehensively and accurately extract all relevant personal information from the provided text and output the results in a structured JSON format.  
If any information in the text is ambiguous or uncertain, the model is asked to explicitly indicate it in the output.  
The output schema and field definitions are provided below.

\subsection*{A.2 Prompt Content}

\begin{tcolorbox}[colback=white,colframe=red!65!yellow,title=Prompt(Chinense),arc=10mm, colbacktitle=red!70!yellow, coltitle=white,fonttitle=\bfseries,breakable]
\begin{Verbatim}[breaklines=true]
你是人物领域的知识抽取专家，******内是你需要进行抽取的文本内容，请你根据所给的文本信息，全面准确地对人物信息进行抽取，并按照JSON格式进行输出，如有模糊或者不确定的地方请说明，输出模板及各字段说明附后。
****** 
人物文本
******
固定输出的JSON格式如下：
{"姓名":"该人物的姓名",
 "别名":"该人物的别名，如字号，别名等",
 "性别":"该人物的性别",
 "民族":"该人物属于哪个民族",
 "所处时代":"该人物所处的时代",
 "籍贯":"该人物的籍贯信息",
 "出生日期":"该人物的出生日期",
 "逝世日期":"该人物的逝世日期",
 "主要成就":[
 {"成就影响":"事件1","发生地点":"地点","发生时间":"时间"},...],
 "主要作品":"个人作品",
 "主要社会关系":[
 {"人物":"人物名字1","关系":"人物名字1与被信息抽取人物的关系，如同僚、上级、下级等"},...],
 "主要家族关系":[
 {"人物":"人物名字1","关系":"人物名字1与被信息抽取人物的关系，如父亲、儿子等"},...],
 "领域":"人物所在领域，如军事家、教育家、思想家等对于个人的概括性描述",
 "历任职务":[
 {"职务1":"职务岗位描述","时间":"担任该职务的开始时间，以年为单位"},
 {"职务2":"职务岗位描述","时间":"担任该职务的开始时间，以年为单位"}]}
\end{Verbatim}
\end{tcolorbox}

\newtcolorbox{myverbatimbox}[1][Prompt(English)]{
	colback=white,
	colframe=teal,
	title=#1,
	arc=10mm,
	colbacktitle=teal,
	coltitle=white,
	fonttitle=\bfseries,
%	drop shadow,
	breakable
}

\begin{myverbatimbox}
\begin{Verbatim}[breaklines=true]
You are an expert in knowledge extraction within the domain of historical and biographical figures. The content enclosed within ****** is the text from which you need to extract information. Based on the provided text, please comprehensively and accurately extract all relevant information about the individual and output the results in JSON format. If there are any ambiguous or uncertain details, please indicate them explicitly. The output template and field descriptions are provided below.
****** 
The text of characters
******
The fixed JSON output format is as follows:
{"Name": "The person's full name",
 "Alias": "Any alternate names, such as courtesy names or pseudonyms",
 "Gender": "The person's gender",
 "Ethnicity": "The ethnic group to which the person belongs",
 "Era": "The historical period or dynasty in which the person lived",
 "BirthPlace": "The person's place of origin",
 "BirthDate": "The person's date of birth",
 "DeathDate": "The person's date of death",
 "MajorAchievements": [
 {"Achievement": "Event 1", "Location": "Place", "Time": "Time"}, ...],
 "MajorWorks": "The person's notable works or publications",
 "MajorSocialRelations": [
 {"Person": "Name of Person 1", "Relation": "Relationship between Person 1 and the extracted individual, e.g., colleague, superior, subordinate"}, ...],
 "MajorFamilyRelations": [
 {"Person": "Name of Person 1", "Relation": "Family relationship between Person 1 and the extracted individual, e.g., father, son"}, ...],
 "Field": "The domain or field the person is known for, e.g., military leader, educator, philosopher, etc.",
 "OfficialPositions": [
 {"Position1": "Description of the post or title", "Time": "The year or period when the person assumed this position"},
 {"Position2": "Description of the post or title", "Time": "The year or period when the person assumed this position"}]}
\end{Verbatim}
\end{myverbatimbox}

\subsection*{A.3 Output Schema Description}

\begin{table}[h]
\begin{tabularx}{\linewidth}{
>{\raggedright\arraybackslash\hsize=0.6\hsize}X
>{\raggedright\arraybackslash\hsize=1.4\hsize}X
}
	\hline
	\textbf{Component} & \textbf{Description (English)}  \\
	\hline
	姓名 & Full name of the individual \\
	别名 & Alternative names, such as courtesy name or pseudonym \\
	性别 & Gender of the individual \\
	民族 & Ethnic group or nationality \\
	所处时代 & Historical era in which the person lived \\
	籍贯 & Place of origin \\
	出生日期 & Date of birth\\
	逝世日期 & Date of death\\
	主要成就 & Major accomplishments and their context\\
	主要作品 & Representative works\\
	主要社会关系 & Non-familial social relations (e.g., colleagues, superiors) \\
	主要家族关系 & Familial relations (e.g., father, son)\\
	领域 & Professional or disciplinary field \\
	历任职务 & Positions held and corresponding time periods  \\
	\hline
\end{tabularx}

\end{table}

\subsection*{A.4 Notes}

\begin{itemize}
	\item For missing or uncertain information, use an empty string (“ ”) or specify as “未知” (unknown).
	\item All field names should remain in Chinese to maintain alignment with the downstream knowledge graph schema.
\end{itemize}

\section*{Appendix B. Huxiang Culture}
\subsection*{B.1 Huxiang Culture}
Huxiang Culture refers to a regional culture form with distinctive features. Dongting Lake and Xiangjiang River Basin in Hunan Province are home to the Huxiang Culture. Huxiang Culture is like the sunlight streaming through the clouds and stars shining in the night sky. “Huxiang” established in the Tang Dynasty has become a vital concept of human geography in Hunan Province. 

\bibliographystyle{elsarticle-num.bst}
\bibliography{Rf.bib}

\end{document}